\documentclass[11pt,letterpaper]{article}

\usepackage{authblk}
\usepackage{emnlp2018}
\usepackage{pslatex}
\usepackage[english]{babel}
\usepackage[utf8]{inputenc}
\usepackage{amsmath}
\usepackage{bm}
\usepackage{graphicx}
\usepackage{tikz}
\usepackage{xcolor}
\usepackage{tabu}
\usepackage{multirow}
\usepackage{url}
\usepackage{rotating}
\usepackage{natbib}
\usepackage{amssymb}
\usepackage{linguex}

\newcommand{\key}[1]{\textbf{#1}}
\newcommand{\soft}[1]{}
\newcommand{\nopreview}[1]{}

\title{What do RNN Language Models Learn about Filler--Gap Dependencies?}

\author[1]{\textbf{Ethan Wilcox}}
\author[2]{\textbf{Roger Levy}}
\author[3,4]{\textbf{Takashi Morita}}
\author[5]{\textbf{Richard Futrell}}

\affil[1]{Department of Linguistics, Harvard University, \tt{wilcoxeg@g.harvard.edu}}
\affil[2]{Department of Brain and Cognitive Sciences, MIT, \tt{rplevy@mit.edu}}
\affil[3]{Primate Research Institute, Kyoto University, \tt{tmorita@alum.mit.edu}}
\affil[4]{Department of Linguistics and Philosophy, MIT}
\affil[5]{Department of Language Science, UC Irvine, \tt{rfutrell@uci.edu}}

\date{}

\begin{document}
\aclfinalcopy
\setlength{\Exlabelwidth}{0.7em}
\setlength{\Exlabelsep}{0.7em}
\setlength{\SubExleftmargin}{1.3em}
\setlength{\Extopsep}{2pt}

\maketitle

\begin{abstract}
RNN language models have achieved state-of-the-art perplexity results and have proven useful in a suite of NLP tasks, but it is as yet unclear what syntactic generalizations they learn. Here we investigate whether state-of-the-art RNN language models represent long-distance \key{filler--gap dependencies} and constraints on them. Examining RNN behavior on experimentally controlled sentences designed to expose filler--gap dependencies, we show that RNNs can represent the relationship in multiple syntactic positions and over large spans of text. Furthermore, we show that RNNs learn a subset of the known restrictions on filler--gap dependencies, known as \key{island constraints}: RNNs show evidence for wh-islands, adjunct islands, and complex NP islands. These studies demonstrates that state-of-the-art RNN models are able to learn and generalize about empty syntactic positions.
\end{abstract}

\section{Introduction}

Many recent advancements in Natural Language Processing have come from the introduction of Recurrent Neural Networks (RNN) \citep{elman1990finding,goldberg2017neural}. One class of RNNs, the Long Short-Term Memory RNN (LSTM) \citep{hochreiter1997long} has been able to achieve impressive results on a suite of NLP tasks, including machine translation, language modeling, and syntactic parsing \citep{sutskever2014sequence,vinyals2015grammar,jozefowicz2016exploring}. But the nature of the representations learned by these models is not properly understood. As these models are being deployed with increasing frequency, this poses both engineering, accountability, and theoretical problems.

One promising line of research aims to crack open these `black boxes' by investigating how LSTM language models perform on  specially controlled sentences designed to draw out behavior that indicates representation of a syntactic dependency.  Using this method, \citet{linzen2016assessing} and \citet{gulordava2018colorless} demonstrated that these models are able to successfully learn the number agreement dependency between a subject and its verb, even when there are intervening elements, and \citet{mccoy2018revisiting} found that RNNs learn the hierarchical rules of English auxiliary inversion. In this paper, we broaden and deepen this line of inquiry by examining what LSTMs learn about an unexplored syntactic relationship: the filler--gap dependency. The filler--gap dependency is novel, insofar as learning it requires the network to generalize about the absence of material. 

For our purposes, \key{filler--gap dependency} refers to a relationship between a \key{filler}, which is a wh-complementizer such as `what' or `who', and a \key{gap}, which is an empty syntactic position licensed by the filler. In example \ref{ex:simple-good}, the filler is `what' and the gap appears after `devoured', indicated with underscores. If the filler were not present, the gap would be ungrammatical, as in \ref{ex:simple-bad}.

\ex. \label{ex:simple} 
\a. I know what the lion devoured \_\_ at sunrise.\label{ex:simple-good}
\b. *I know that the lion devoured \_\_ at sunrise.\label{ex:simple-bad}

There is also a semantic relationship between the filler and the gap, in the sense that ``what'' is semantically the direct object of ``devoured''. In this work, we study the behavior of language models, and so we treat the filler--gap dependency purely as a licensing relationship. 

\citet{elman1991distributed} found that simple distributed models have some success predicting post-verbal gaps in  sentences containing object-extracted relative clauses. However, correct representation of filler--gap dependencies and the constraints on them has proven challenging even in hand-engineered symbolic models. Furthermore, they are subject to numerous complex  \key{island constraints} \citep{ross1967constraints}. Because of their complexity and ubiquity, these dependencies have figured prominently in arguments that natural language would be unlearnable by children without a great deal of innate knowledge \citep{phillips2013nature} \citep[cf.][]{pearl2013syntactic, ellefson2000subjacency}

The remainder of the paper is structured as follows. Section~\ref{sec:methods} presents our methods in more detail. Section~\ref{sec:basic} gives evidence that LSTM language models represent the basic filler--gap dependency in multiple syntactic positions despite intervening material. Section~\ref{sec:islands} investigates whether LSTM language models are sensitive to various constraints: wh-islands, adjunct islands, complex NP islands, and subject islands. We find that the language models are sensitive to some but not all of these constraints. Section~\ref{sec:conclusion} concludes.

\section{Methods}
\label{sec:methods}

\subsection{Language models}

We study the behavior of two pre-existing LSTMs trained on a language modeling objective over English text. Our first model is presented in \citet{jozefowicz2016exploring} under the name \textit{BIG LSTM+CNN Inputs}; we call it the \key{Google model}. It was trained on the One Billion Word Benchmark \citep{chelba2013one} and has two hidden layers with 8196 units each. It uses the output of a character-level Convolutional Neural Network (CNN) as input to the LSTM. This model has the best published perplexity for English text. Our second model is the one presented in the supplementary materials of \citet{gulordava2018colorless}, which we call the \key{Gulordava model}. Trained on 90 million tokens of English Wikipedia, it has two hidden layers of 650 units each. Our goal in using these models is to provide two samples of the state-of-the-art. As a baseline, we also study an $n$-gram model trained on the One Billion Word Benchmark (a 5-gram model with modified Kneser-Ney interpolation, fit by KenLM with default parameters) \citep{heafield2013scalable}. 

\subsection{Dependent variable: Surprisal}

We investigate RNN behavior primarily by studying the \key{surprisal values} that an RNN assigns to words and sentences. Surprisal is log inverse probability:
\begin{equation*}
S(x_i) = -\log_2 p(x_i|h_{i-1}),
\end{equation*}
where $x_i$ is the current word or character, $h_{i-1}$ is the RNN's hidden state before consuming $x_i$, and the probability is calculated from the RNN's softmax activation. The logarithm is taken in base 2, so that surprisal is measured in bits.

The degree of surprisal for a word or sentence tells us the extent to which that word or sentence is unexpected under the language model's probability distribution. It is known to correlate directly with human sentence processing difficulty \citep{hale2001probabilistic,levy2008expectation,smith2013effect}. In this paper, we look for cases where the surprisal associated with an an unusual construction---such as a gap---is ameliorated by the presence of a licensor, such as a wh-word. If the models learn that syntactic gaps require licensing, then sentences with licensors should exhibit lower surprisal than minimally different pairs that lack a proper licensor.

\subsection{Experimental design} \label{sec:licensin-description}

We test whether the LSTM language models have learned filler--gap dependencies by looking for a 2x2 interaction between the presence of a gap and the presence of a wh-licensor. This interaction indicates the extent to which a wh-licensor reduces the surprisal associated with a gap, so we call it the \key{wh-licensing interaction}. In studying constraints on filler--gap dependencies, we look for interactions between the wh-licensing interaction and other factors: for example, whether the wh-licensing interaction decreases when a gap is in a syntactic island position as opposed to a syntactically licit position (Section~\ref{sec:islands}).

We use experimental items where the gap is located in an obligatory argument position, e.g. in subject position or as the direct object of a transitive verb, as judged by the authors. The phrase with the gap is embedded inside a complement clause. We chose this paradigm over bare wh-questions because it eliminates do-support and tense manipulation of the main verb, resulting in higher similarity across conditions. Each item appears in four conditions, reflecting a $2 \times 2$ experimental design manipulating presence of a wh-licensor and presence of a gap. For example:\footnote{We indicate the gap position with underscores for expository purposes, but these underscores were not included in experimental items.}

\ex. \label{ex:li-overview}
\a. I know that the lion devoured a gazelle at sunrise. [no wh-licensor, no gap]\label{ex:li-that-nogap}
\b. *I know what the lion devoured a gazelle at sunrise. [wh-licensor, no gap]\label{ex:li-wh-nogap}
\c. *I know that the lion devoured \_\_ at sunrise. [no wh-licensor, gap]\label{ex:li-that-gap}
\d. I know what the lion devoured \_\_ at sunrise. [wh-licensor, gap]\label{ex:li-wh-gap}

We measure surprisal in two places: at the word immediately following a (filled) gap and summed over the whole region from the gap to the end of the embedded clause. We look at immediate-word surprisal because a gap's licitness should have local effects on network expectation. 
We look at whole-region surprisal because the presence of a filler also changes expectations about overall well-formedness of the sentence---a global phenomenon. Until the final punctuation is reached in \ref{ex:li-wh-nogap} there are potential gap-containing continuations that render the sentence syntactically licit (e.g. `with \_\_.'). Therefore, we might expect no large spike in surprisal at any one point, but small increases in surprisal when the network encounters filled argument-structure roles and at the end of the sentence. Measuring summed surprisal captures these distributed, global effects. 

If the network is learning the licensing relationship between fillers and gaps then two things should be true: First, if a wh-licensor sets up a global expectation for the presence of a gap, then in sentences containing a wh-licensor but no gap we expect higher surprisal in syntactic positions where a gap is likely to occur resulting in higher summed surprisal. That is, $S(\ref{ex:li-wh-nogap}) - S(\ref{ex:li-that-nogap})$ should be a large positive number. Second, the presence of a gap in the absence of a wh-licensor should also result in higher surprisal than when the wh-licensor is present, that is $S(\ref{ex:li-wh-gap}) - S(\ref{ex:li-that-gap})$ should be a large negative number. Given the four sentences in \ref{ex:li-overview}, the full wh-licensing interaction is: (S\ref{ex:li-wh-nogap} - S\ref{ex:li-that-nogap}) - (S\ref{ex:li-wh-gap} - S\ref{ex:li-that-gap}) This represents how well the network learns both parts of the licensing relationship. A positive wh-licensing interaction means the model represents a filler-gap dependency between the wh-word and the gap site; a licensing interaction indistinguishable from zero indicates no such dependency. For the purposes of brevity, we will give examples that mirror item \ref{ex:li-wh-gap}, above, but items of type \ref{ex:li-that-nogap}--\ref{ex:li-that-gap} were also constructed in order to calculate the full licensing interaction.

Following standard practice in psycholinguistics, we derive the statistical significance of the interaction from a mixed-effects linear regression model predicting surprisal given sum-coded conditions \citep{baayen2008mixed}. We include random intercepts by item; random slopes are not necessary because we do not have repeated observations within items and conditions \citep{barr2013random}. In our figures, error bars represent 95\% confidence intervals of the contrasts between conditions, computed by subtracting out the by-item means before calculating the intervals as advocated in \citet{masson2003using}. \footnote{Our studies were preregistered on \url{aspredicted.org}: To see the preregistrations go to \url{aspredicted.org/}$X$\url{.pdf} where $X \in \{\texttt{md5ax}, \texttt{hd2df}, \texttt{mp9dv}, \texttt{uu8b5}, \texttt{rj2sk}\}$.}

Although our method can indicate whether there is a link between fillers and gaps, the relationship between language model probability and grammaticality is complex \citep{lau2017grammaticality} and interpreting our patterns in terms of grammaticality judgments would require auxiliary assumptions that we don't pursue here. To be clear: our goal is to investigate whether RNNs model the probabilistic dependencies between fillers and gaps \textit{at all}, not whether the outputs of such models can be used to classify sentences as `grammatical' or not. 

\section{Representation of filler--gap dependencies}
\label{sec:basic}
The filler--gap dependency has three basic characteristics. First, the relationship is \key{flexible}: wh-phrases can license gaps in diverse syntactic positions. Second, the relationship is \key{robust to intervening material}: syntactic position, not linear distance, determines grammaticality. Third, the relationship is \key{one-to-one}: except in certain special cases, one wh-phrase licenses one gap. In this section, we demonstrate that the RNNs have learned these three properties of filler--gap dependencies by comparing their performance to a simple $n$-gram baseline model.

\subsection{Flexibility of Wh-Licensing}
If the RNN has learned the flexibility of the filler--gap dependency, then we predict to find a wh-licensing interaction when the gap appears in subject, object, and indirect object positions: 

\ex. \label{ex:gap-pos}
\a. I know who \_\_ showed the presentation to the visitors yesterday. [\emph{subj}] \label{ex:gap-pos-subj}
\b. I know what the businessman showed \_\_ to the visitors yesterday. [\emph{obj}] \label{ex:gap-pos-obj}
\c. I know who the businessman showed the presentation to \_\_ yesterday. [\emph{pp}] \label{ex:gap-pos-pp}

To test the flexibility of the model's filler--gap dependency representation, we created 21 test items containing either an obligatorily ditransitive verb, or a transitive verb with an obligatorily argument-taking preposition, as in \ref{ex:gap-pos}. The obligatoriness of verb and preposition transitivity was judged by the authors. To control for the infrequent \emph{wh-licensor}--\emph{verb} bigram when the gap is in subject position, in all cases the embedded clause was separated from the wh-phrase by either an adverbial (e.g. ``despite protocol'') or by words introducing a secondary embedded clause (e.g. ``my brother said''). For each item, we created three variants: \emph{subj}, \emph{obj}, and \emph{pp}, corresponding to the items in Example \ref{ex:gap-pos}.

\begin{figure*}[ht!]
\begin{minipage}{0.24\textwidth}
(a)
\includegraphics[width=\textwidth]{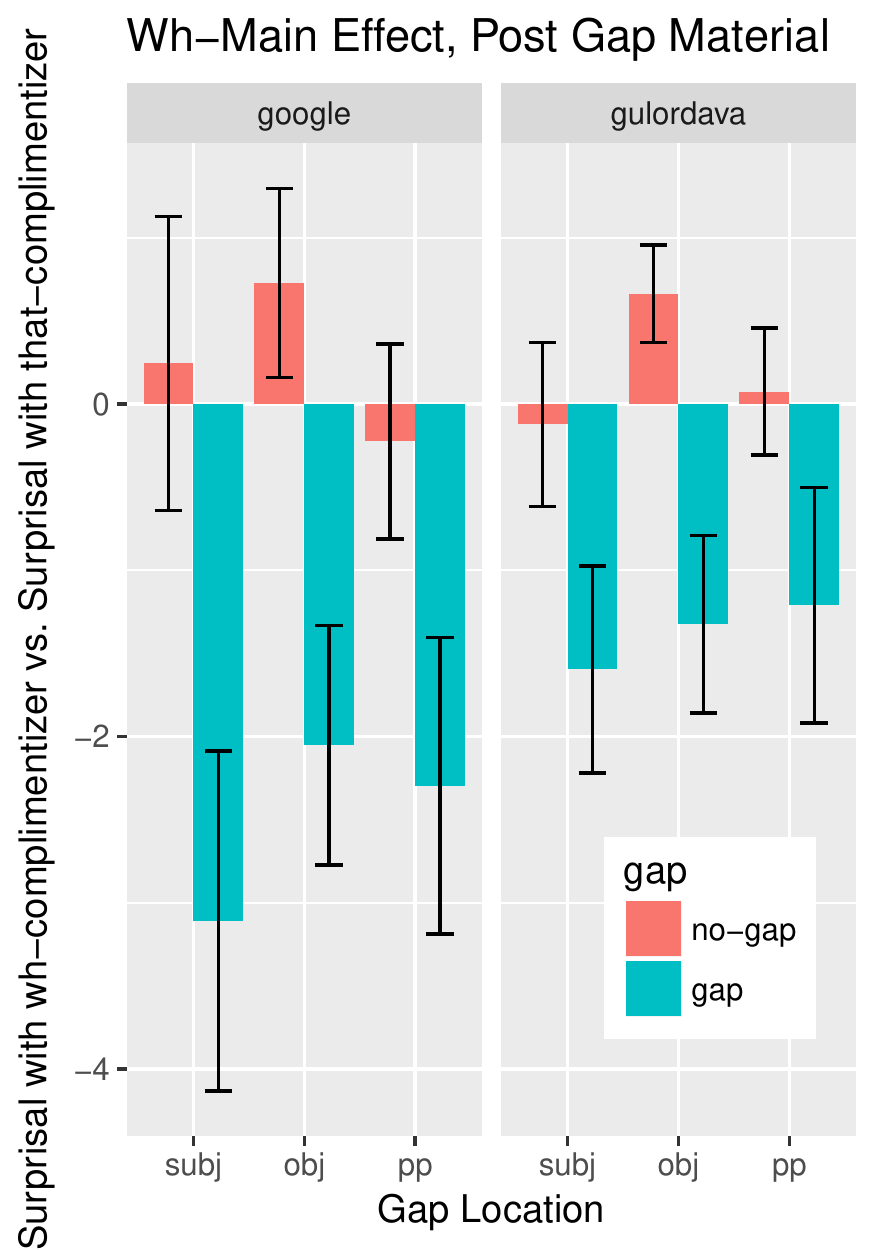}
\end{minipage}
\begin{minipage}{0.24\textwidth}
(b)
\includegraphics[width=\textwidth]{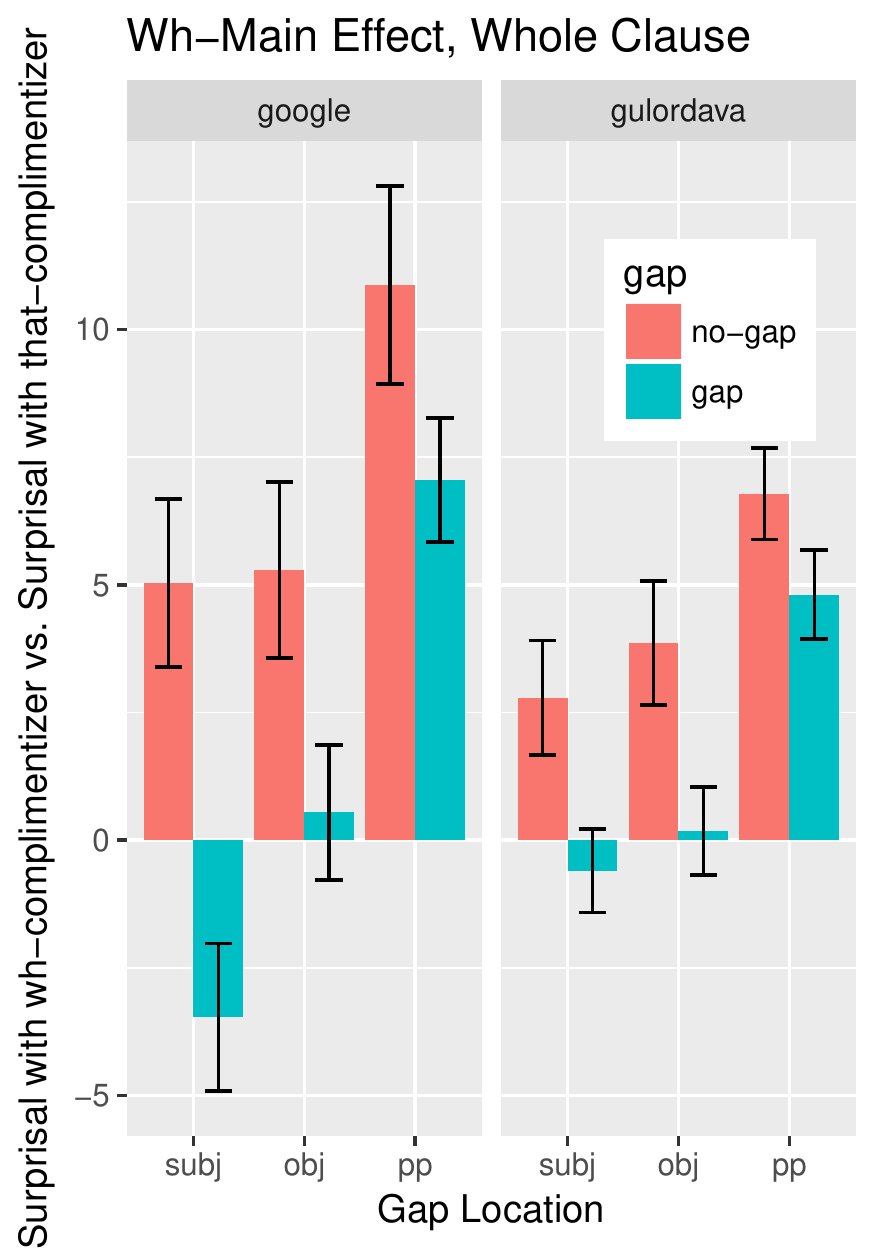}
\end{minipage}
\begin{minipage}{0.24\textwidth}
(c)
\includegraphics[width=\textwidth]{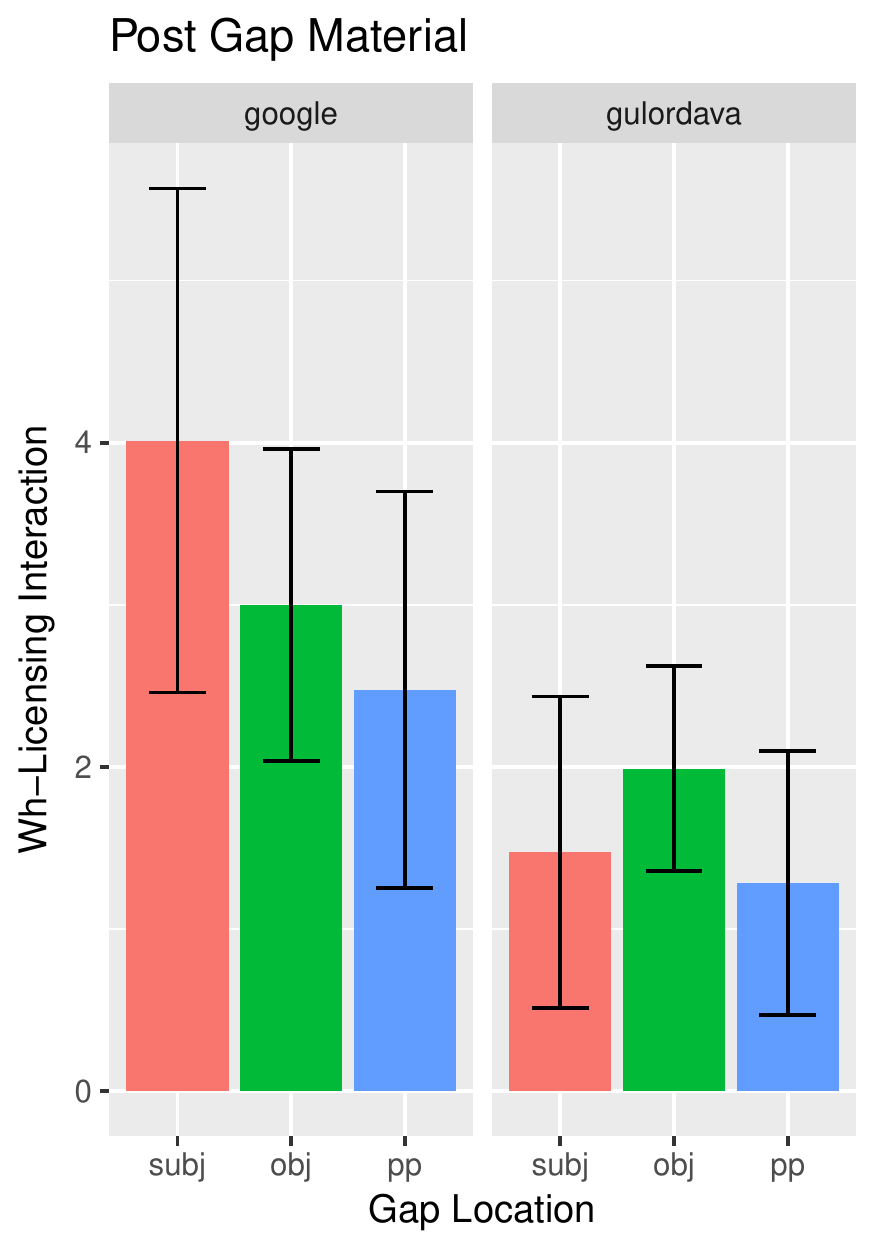}
\end{minipage}
\begin{minipage}{0.24\textwidth}
(d)
\includegraphics[width=\textwidth]{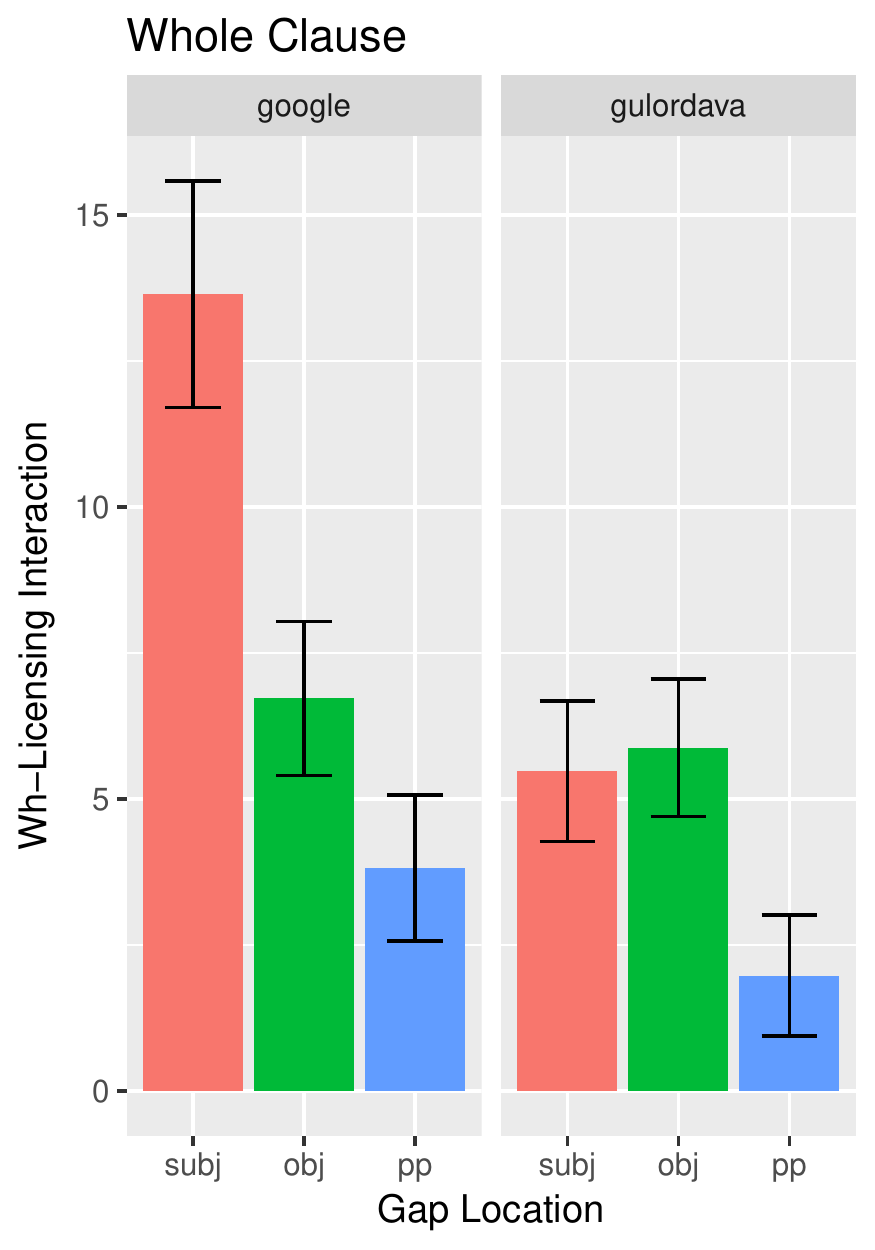}
\end{minipage}\\
\begin{minipage}{0.98\textwidth}
\includegraphics[width=1\textwidth]{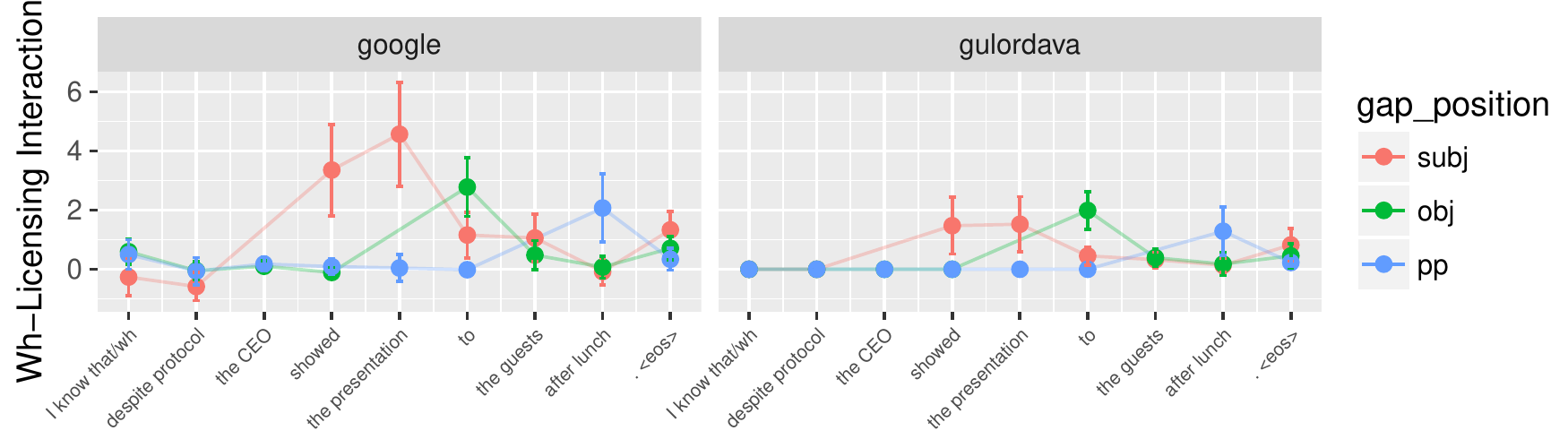}
\end{minipage}
\caption{Wh-licensing by syntactic position. Charts (a) and (b) show the effect of wh-licensors on surprisal; (c) and (d) show the wh-licensing interaction by syntactic position. The difference between the non-gapped and gapped conditions (red and blue bars) in (a) and (b) correspond to the total licensing interaction, or the height of the bars in (c) and (d). The bottom chart displays wh-licensing interaction summed across all words within each region.}
\label{fig:position}
\end{figure*}

The top row of Figure \ref{fig:position} demonstrates how the wh-licensing interaction was calculated for this experiment. The two panels at left show the main effect of wh-licensing, with surprisal in post-gap material shown in (a) and summed whole-clause surprisal in (b). The red bars indicate the effect of a wh-licensor on surprisal in the non-gapped condition, or $S$\ref{ex:li-wh-nogap}--$S$\ref{ex:li-that-nogap}, to use the example from \ref{sec:licensin-description}. The blue bars show the effect of a wh-licensor on surprisal in the gapped conditions, or $S$\ref{ex:li-wh-gap}--$S$\ref{ex:li-that-gap}, to use the same example. The difference between the red bars and the blue bars in each condition is the licensing interaction, which is shown directly in (c) and (d). Not pictured are results from the $n$-gram baseline model, which yielded exactly 0 licensing interaction in all positions. 

The bottom row of Figure \ref{fig:position} shows a region-by-region visualization of wh-licensing interaction. Region-by-region behavior is consistent across conditions: The licensing interaction spikes in the immediate post-gap material and returns to near zero levels for the rest of the sentence. The height of the licensing `spike' in each condition is equivalent to the size of the wh-licensing interaction in (c), and the difference between the bars in (a). Meanwhile, the area under the `wh-licensing curve' is equivalent to the summed wh-licensing interaction shown in (d) and the difference between the bars in (b). All of these wh-licensing interactions are significant ($p<0.001$ in all cases).

This experiment was designed to test whether licensing interaction exists in multiple syntactic positions, which we turn to now. In the post-gap material, there is no significant difference in licensing interaction between conditions. But when we sum wh-licensing interaction across the entire embedded clause model behavior does diverge. For the Gulordava model, there is no significant difference between the three variants. For the Google model there is a significant reduction in licensing effect between the \textit{subj} and \textit{obj} variants ($p < 0.01$) and the \textit{subj} and \textit{pp} variants ($p < 0.001$). The stronger licensing effects for subject gaps indicates that the networks have a stronger expectation for gaps in this position. This matches human online processing results, in so far as gap expectation may be one reason why subject-extracted clauses are easier to process than other clauses \citep{king1991individual}. Overall, these experiments provide strong evidence that both models are learning the filler--gap dependency. Furthermore, both RNN models are learning the flexibility of the dependency, as they exhibit similar wh-licensing effects for all three argument roles tested.


\subsection{Robustness of Wh-Licensing to Intervening Material}

All syntactic dependencies are robust to intervening material. In \ref{ex:length}, the dependency is determined by the syntactic relationship between the complementizer `what' and the position of the gap; modifying the subject doesn’t change the relationship, and thus has no effect on filler--gap licensing:

\ex. \label{ex:length}
\a. I know what your friend gave \_\_ to Sam during the picnic yesterday. \label{ex:length-short}
\b. I know what your new friend from the south of France who only just arrived last week gave \_\_ to Sam during the picnic yesterday. \label{ex:length-long} 

Having shown previously that RNNs have expectations for filler--gap dependencies, in this section we ask how well they are able to maintain those expectations over intervening material. We designed 21 sentences, like those in \ref{ex:length}, with an obligatorily transitive verb and either an indirect object or a PP modifier. For each sentence we produced four variants, a short-modified version with 3-5 extra intervening words between the wh-licensor and the gap site, a medium version with 6-8 additional words and a long version, with 8-12 additional words. In all cases the extra material modified the subject of the embedded clause. For each length gradation we produced two further variants: one in which the direct object was extracted (\emph{obj}, as in \ref{ex:length}) and one variant in which the indirect object or prepositional object was extracted (\emph{goal}, where `Sam' is in \ref{ex:length}). For each variant, we measured the wh-licensing interaction in the post-gap material and across the embedded clause. Treating the number of intervening words as a continuous variable, we calculated the correlation between the length of the intervener and the strength of the wh-licensing interaction. Optimally we would find zero correlation; a negative correlation indicates that the strength of the interaction decays with increasing intervening words.

Results of this study can be seen in Figure \ref{fig:length}. First, as a baseline, across the eight experiments shown below, the average number of positive licensing interaction measurements was 86.4\%. The vast majority of the time, the presence of both a filler and a gap reduced surprisal superadditively, producing a positive licensing interaction. Moving on to the effect of intervener length itself: For the Google model, intervener length was not a significant  predictor of wh-licensing interaction in any of the conditions. For the Gulordava model, intervener length was not a significant predictor of wh-licensing interaction size when measurements were taken across the entire embedded clause. But length did correlate with wh-licensing interaction size when measured in the post-gap material for the object position ($\beta=0.0289, p=0.0219$) and goal position ($\beta=0.0047, p=0.0432$). These extremely small effect sizes, combined with the otherwise mixed results from both models, indicate that interveners do not consistently attenuate the size of the licensing interaction. 

While inconsistent with the formal linguistic literature on filler--gap dependencies, the negative values of all but one of the correlations are consistent with known effects in human sentence processing, where increasing distance between fillers and gaps usually causes processing slowdown \cite{grodner2005consequences,bartek2011search}. In the n-gram baseline, all licensing effects are exactly zero, indicating the $n$-gram model has no representation of the filler--gap dependency.

\begin{figure}
\newcommand{\mytextwidth}{0.22\textwidth}
\begin{tabular}{cc}
\includegraphics[width=\mytextwidth]{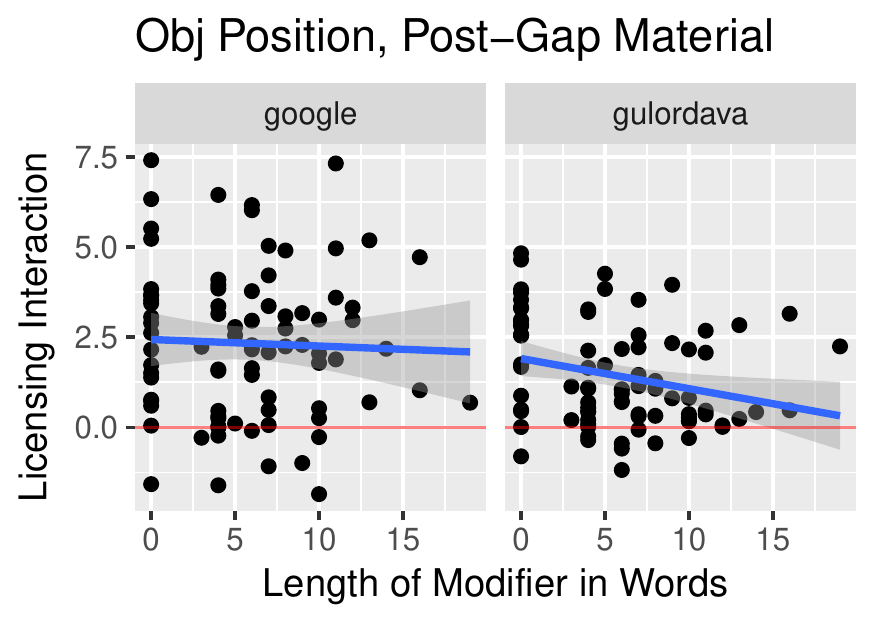}
&
\includegraphics[width=\mytextwidth]{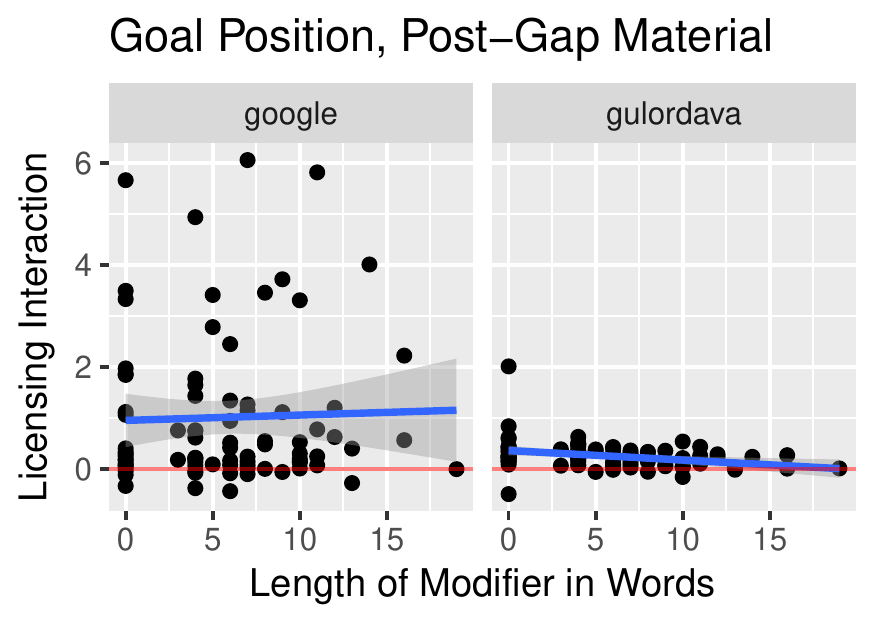}
\\
\includegraphics[width=\mytextwidth]{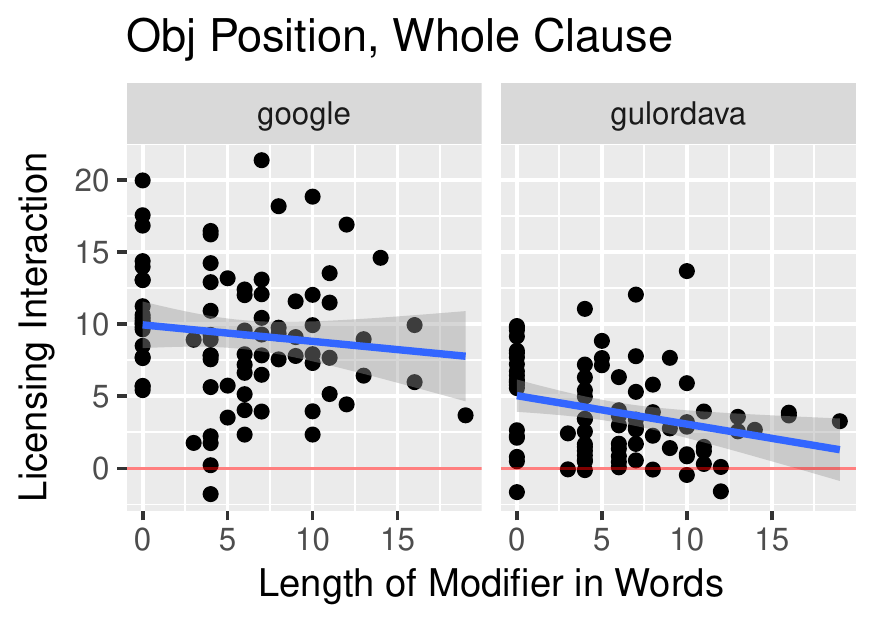}
&
\includegraphics[width=\mytextwidth]{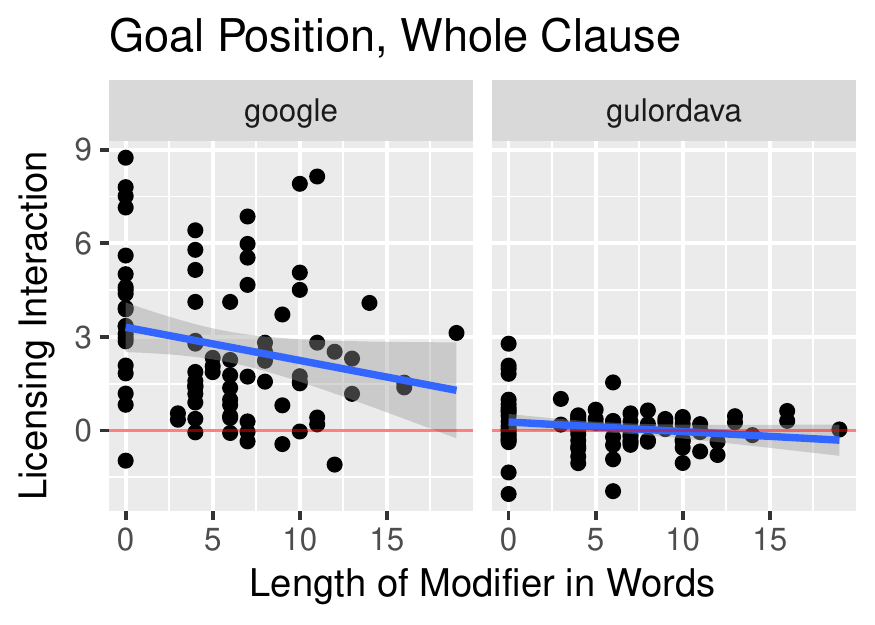}
\end{tabular}
\caption{Wh-licensing interaction as a function of intervener length. Zero is marked with a red line.}
\label{fig:length}
\end{figure}

\subsection{Multiple Gaps}
Except for a few special cases, such as with across-the-board (ATB) movement and parasitic gaps, a one-to-one relationship must be maintained between the wh-phrase and the gap it licenses. The presence of two gaps in \ref{ex:doublegap-illicit} violates this one-to-one relationship, accounting for its relative badness compared to \ref{ex:doublegap-licit} and \ref{ex:doublegap-licit-2}. 

\ex. \label{ex:doublegap}
\a. I know what the lion devoured \_\_ at sunrise.\label{ex:doublegap-licit}
\b. I know what \_\_ devoured a mouse at sunrise.\label{ex:doublegap-licit-2}
\c. * I know what \_\_ devoured \_\_ at sunrise.\label{ex:doublegap-illicit} 

To test whether RNNs have learned this one-to-one feature of wh-licensing, we created 21 items all with gaps in object position like those in \ref{ex:doublegap}, with two variants: one without a subject gap like \ref{ex:doublegap-licit} (\textit{no-subj-gap}) and one with a subject gap, as in \ref{ex:doublegap-illicit} (\textit{subj-gap}). We took special care to use only obligatorily transitive verbs. Half of the test items contained `what' and half `who' as wh-licensors. We measured the wh-licensing interaction for the two RNN models and the $n$-gram model, in both the post-gap PP and across the embedded phrase.

\begin{figure}
\begin{minipage}{0.22\textwidth}
\includegraphics[width=\textwidth]{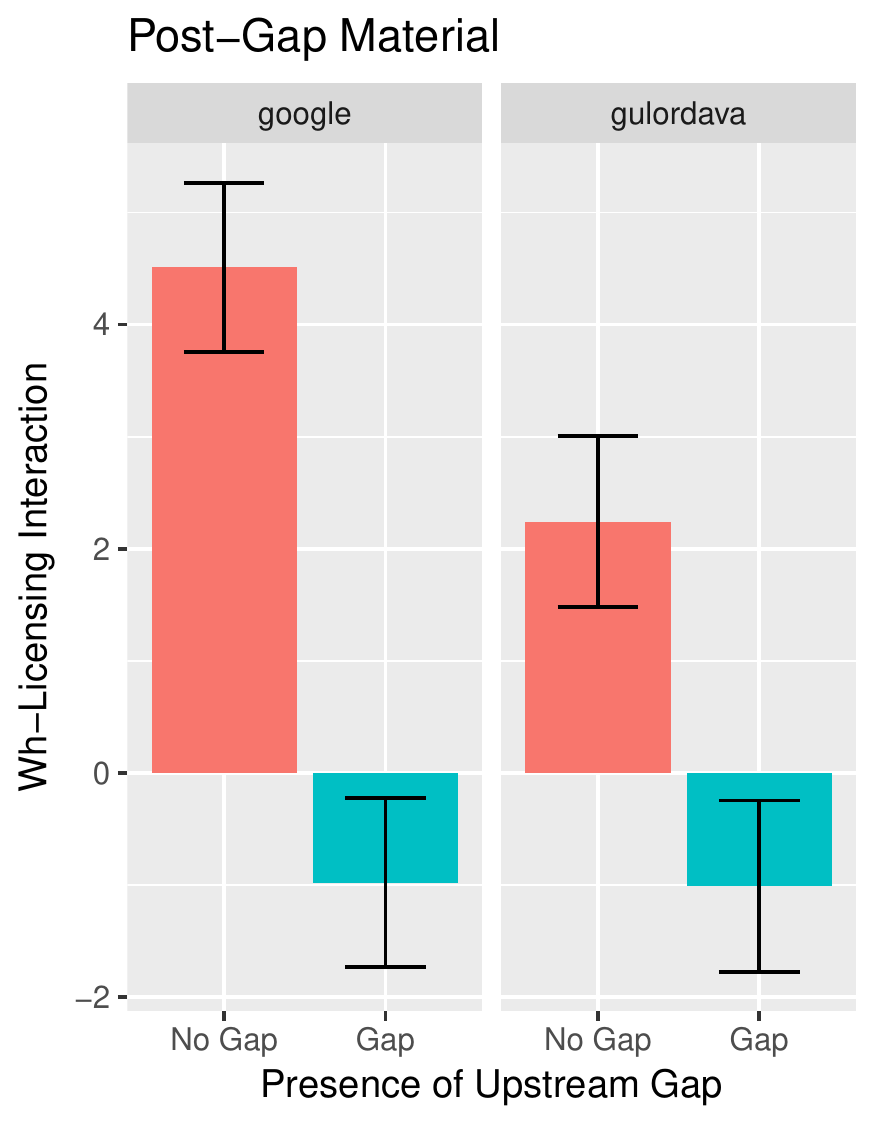}
\end{minipage}
\begin{minipage}{0.22\textwidth}
\includegraphics[width=\textwidth]{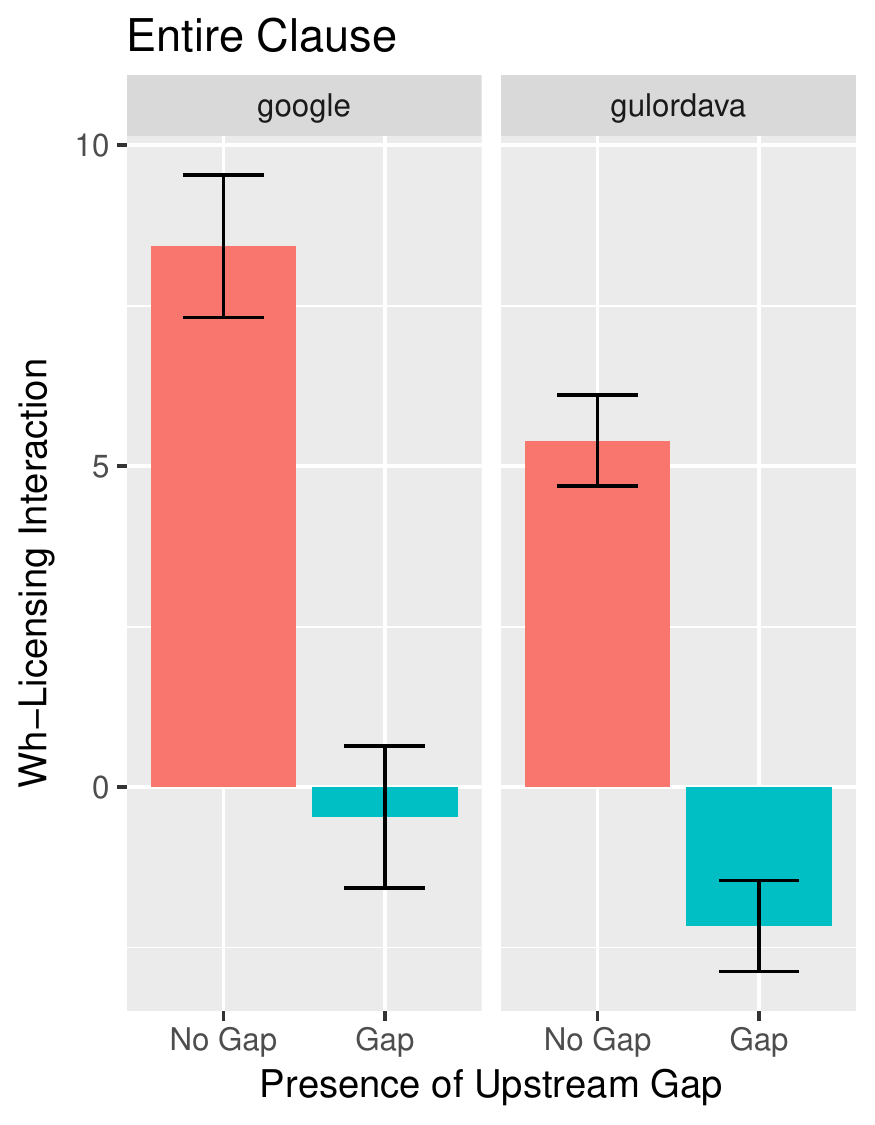}
\end{minipage}
\caption{Wh-Licensing Interaction as a function of Double Gapping: Singly-gapped sentences are shown in red, doubly-gapped sentences in blue. Prepositional Phrases following the gap constitute post-gap material.}
\label{fig:doublegap}
\end{figure}

Figure \ref{fig:doublegap} shows the results of this experiment. First, the relatively high bars in the grammatical \textit{no-subject-gap} condition is another example of the RNN learning the filler--gap dependency; the $n$-gram baseline (not shown) exhibits no wh-licensing interaction under this condition. For the two LSTMs, the presence of an upstream gap increases surprisal in the target region, resulting in a significantly lower licensing effect across the board ($p < 0.001$ in all conditions). Meanwhile, the presence of a gap in the baseline condition results in no significant change in wh-licensing interaction. Overall these experiments demonstrate that the LSTMs have learned the last of the three main filler--gap dependency characteristics, and---for the typical object position---expect wh-phrases to be paired with only one gap.

\section{Syntactic islands}
\label{sec:islands}

Even though the filler--gap dependency is flexible and potentially unbounded, it is not entirely unconstrained. \citet{ross1967constraints} identified five syntactic positions in which gaps are illicit, dubbing them \key{syntactic islands}. It remains an open question  whether these ``island constraints'' are true grammatical constraints, or whether they are effects of processing difficulty or discourse-structural factors \citep{ambridge2008island,hofmeister2010cognitive,sprouse2014experimental}. 

In the following experiments, we examine whether RNN language models have learned constraints on filler--gap dependencies by comparing the wh-licensing interaction in non-islands to that within islands. The strongest evidence for an island constraint would be if the wh-licensing interaction goes to zero for a gap in island position, implying that, in the distribution over strings implied by the network, the appearance of a wh-licensor is totally unrelated to the appearance of a gap in the island position. More generally, we can look for a weakened wh-licensing interaction for island vs. non-island positions, which would mean that the network believes a relationship between the wh-licensor and the island gap is less likely. A positive but nonzero wh-licensing interaction would be in line with human acceptability judgments, which do not always categorically rule out gaps in island positions \citep{ambridge2008island}, and with human online processing experiments, which have shown that gap expectation is attenuated during processing of areas where gaps cannot occur licitly, but does not always disappear entirely \citep{stowe1986parsing,traxler1996plausibility,phillips2006real}. Therefore, in this section we take a significant reduction in the island relative to the non-island case to constitute evidence that the model has `learned' the constraint.


\subsection{Wh-Island Constraint}

A gap cannot appear inside doubly nested clauses headed by wh-complementizers. This phenomenon is called the \key{Wh-Island Constraint} (WHC). \ref{ex:wh} gives three sentences that demonstrate this phenomenon. As these three sentence variants will serve as the basis for our experiment we give each variant a condition name, on the top, and a brief description below. We will use this three-row expository technique---name, example, description---for each of the island conditions tested in this section and use condition names to label graphs and figures.

\ex. \label{ex:wh}
\small{}
\a. \label{ex:wh-noisland}
	\begin{tabular}{ p{6.5cm} }
    \textit{null-comp} \\ 
    \textbf{I know what Alex said your friend devoured \_\_ at the party.}\\
	Extraction from the object position of an embedded clause with a null complementizer. No island violations. \\
	\end{tabular}
\b. \label{ex:wh-noisland-that}
	\begin{tabular}{ p{6.5cm} } 
    \hline
	\textit{that-comp} \\ 
	\textbf{I know what Alex said that your friend devoured \_\_ at the party.}\\
	Extraction from an embedded clause headed with the complementizer “that.” No island violations. \\
    \end{tabular}
\c. \label{ex:wh-island}
	\begin{tabular}{ p{6.5cm} } 
    \hline
	\textit{wh-comp} \\ 
	\textbf{*I know what Alex said whether your friend devoured \_\_ at the party.}\\
	Extraction from an embedded clause headed with the complementizer “whether.” WHC violation. \\
	\end{tabular}

To test whether our LSTM language models have learned this constraint, we constructed 24 items following the conditions in \ref{ex:wh}. We measured the wh-licensing interactions at the sentence final PP, as well as across the entire embedded clause for both conditions.

\begin{figure}
\begin{minipage}{0.22\textwidth}
\includegraphics[width=\textwidth]{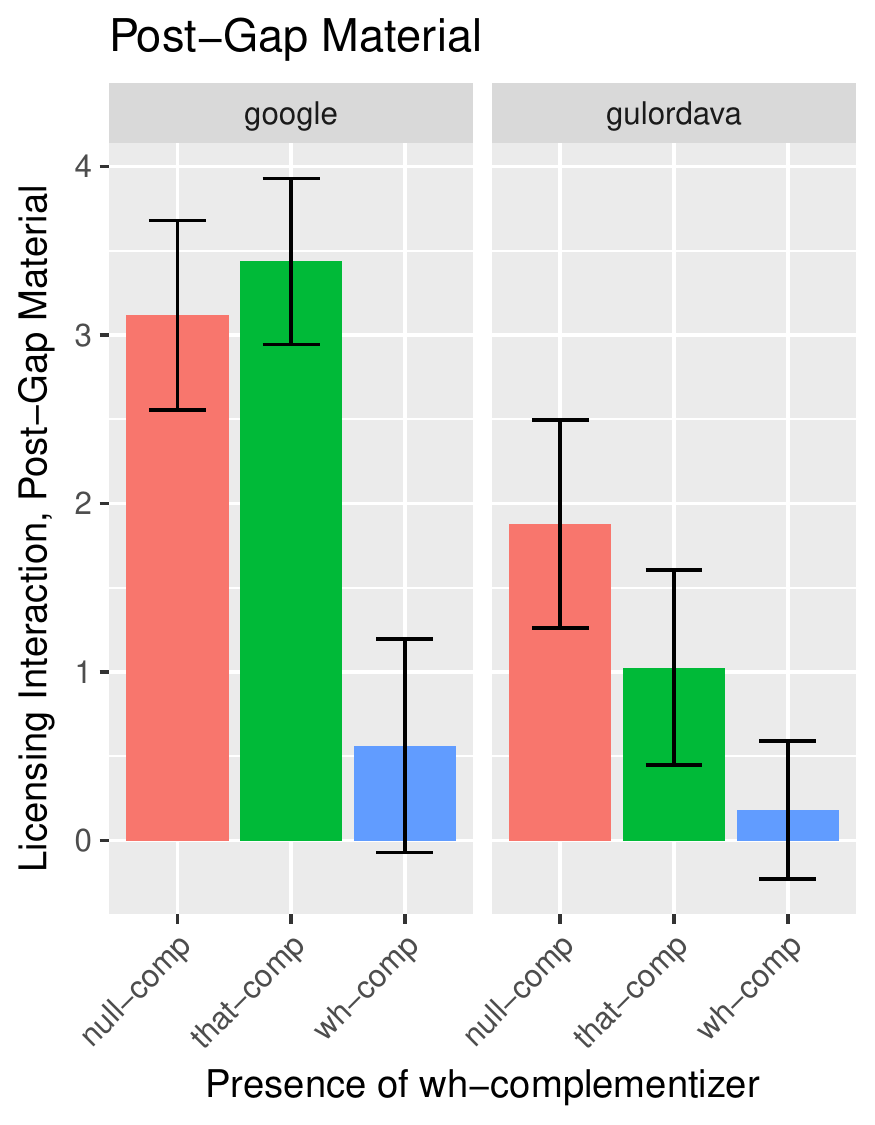}
\end{minipage}
\begin{minipage}{0.22\textwidth}
\includegraphics[width=\textwidth]{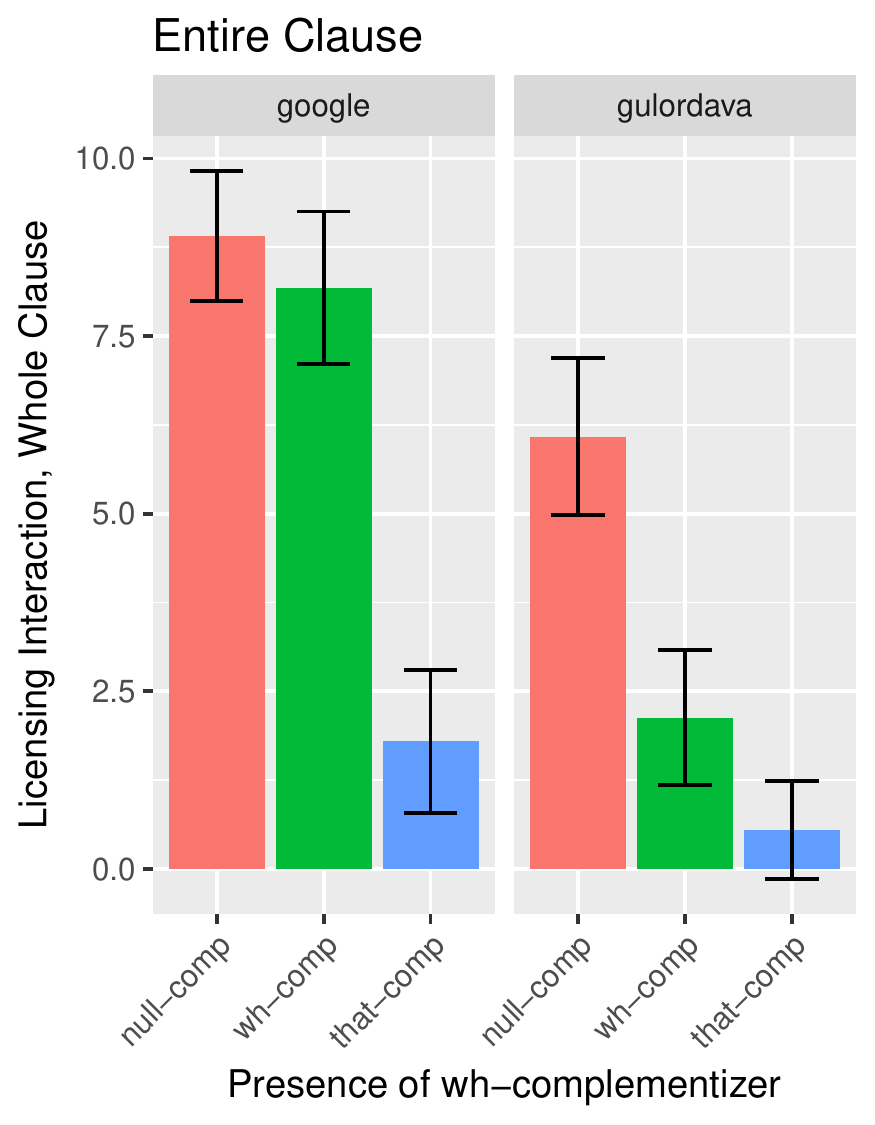}
\end{minipage}
\caption{Effect of embedded clause complementizer on wh-licensing interaction. Post-gap material effect is in the left panel, whole-clause effect on the right panel.}
\label{fig:wh-island}
\end{figure}

Figure \ref{fig:wh-island} shows the wh-licensing interaction for both LSTMs, with non-island conditions in red and green and island conditions in blue. In all conditions, extraction out of a wh-island resulted in a significantly lower licensing interaction than extraction out of a null-headed embedded clause ($p < 0.01$). For the Google model, extraction out of an island resulted in significantly lower wh-licensing interaction than extraction out of a that-headed embedded clause ($p<0.001$), and while the Gulordava model showed similar behavior, none of the reductions were significant ($p=0.071$ for the post gap material and $p=0.052$ for the whole clause measurement).  In all cases there was no significant difference between extraction out of the two non-island conditions, except for in the Gulordava model whole-clause condition, where licensing interaction for the \textit{that-comp} condition was significantly lower than the \textit{null-comp} condition ($p < 0.001$). These results indicate that the Google model has learned the wh-island constraint insofar as it has relatively similar expectations for extraction from null-headed and that-headed clauses, which differ from from its expectations about wh-headed clauses. The Gulordava model has learned wh-islands, but gradiently, treating that-headed embedded clauses as a semi-island condition.

\subsection{Adjunct Island Constraint}

Gaps cannot be licensed in an adjunct clause, as demonstrated by the relative unacceptability of \ref{ex:adj-back} and \ref{ex:adj-front}, compared to \ref{ex:adj-obj}. We will refer to this constraint as the \key{Adjunct Constraint} (AC).

\ex. \label{ex:adjunct}
\small{}
\a. \label{ex:adj-obj}
	\def\arraystretch{1.1}
	\begin{tabular}{ p{6.5cm} }
    \textit{object} \\ 
    \textbf{I know what the librarian in the dark blue glasses placed \_\_ on the wrong shelf.}\\
	\small{Material is extracted from the object position of the embedded verb.  No island violations.} \\
	\end{tabular}
\b. \label{ex:adj-back}
	\begin{tabular}{ p{6.5cm} } 
    \hline
	\textit{adjunct-back} \\ 
	\textbf{*I know what the patron got mad after the librarian placed \_\_ on the wrong shelf.}\\
	\small{Material is moved from the object position of an embedded sentential adjunct. AC violation.} \\
	\end{tabular}
\c. \label{ex:adj-front}
	\begin{tabular}{ p{6.5cm} } 
    \hline
	\textit{adjunct-front} \\
	\textbf{*I know what, after the librarian placed \_\_ on the wrong shelf, the patron got mad.}\\ 
	\small{Material is moved from an embedded sentential adjunct that has been fronted to before the main verb of the embedded clause. AC violation.} \\
	\end{tabular}

To test whether RNNs were sensitive to the AC we devised 20 items following the variants in \ref{ex:adjunct}. Filler material was added to the \textit{object} condition to control for sentence length across variants. We used three different prepositions to construct temporal adjuncts: `while', `after' and `before'. We measured the wh-licensing interaction in the post-gap PP and across the entire embedded clause.

\begin{figure}
\begin{minipage}{0.22\textwidth}
\includegraphics[width=\textwidth]{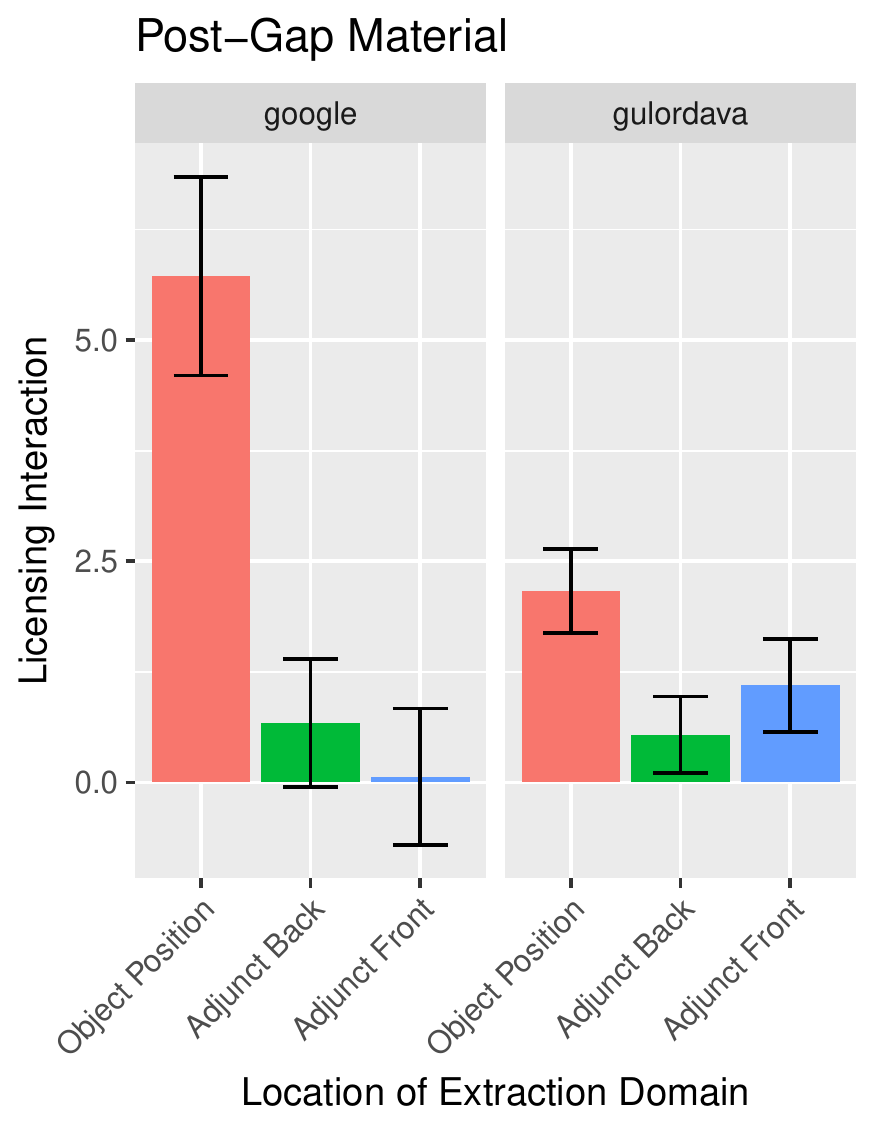}
\end{minipage}
\begin{minipage}{0.22\textwidth}
\includegraphics[width=\textwidth]{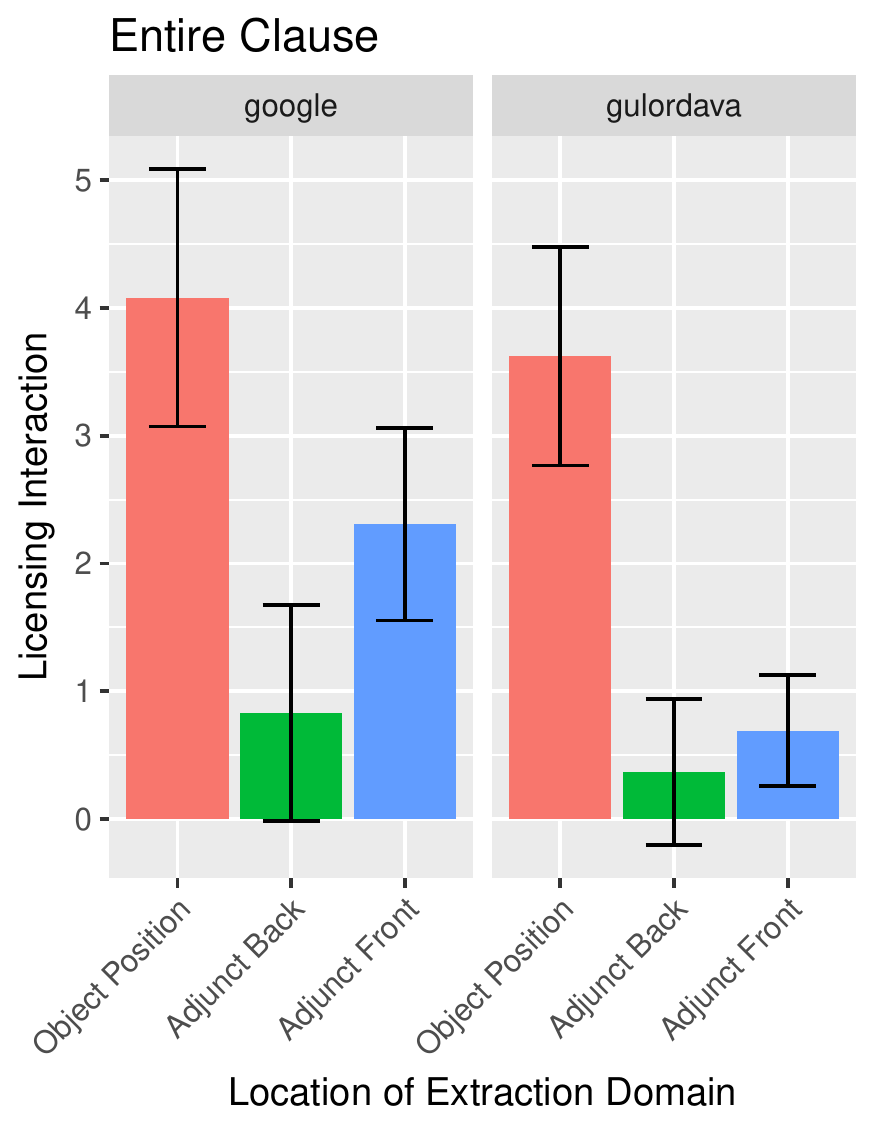}
\end{minipage}
\caption{Effect of extraction site on wh-licensing interaction for adjunct islands. Post-gap material effect is in the left panel, whole-clause effect on the right panel.}
\label{fig:adjunct-results}
\end{figure}

Figure \ref{fig:adjunct-results} shows the wh-licensing interaction for both models. For the Google model there is a significant ($p<0.001$) reduction in wh-licensing interaction between the \textit{object} condition and the two adjunct conditions when measurement is taken in the post-gap material. The difference in licensing is also significant when measurements are taken across the  embedded clause ($p<0.05$ for the \textit{object}--\textit{adj-front} difference and $p<0.01$ for the \textit{object}--\textit{adj-back} difference). The Gulordava model shows similar results. In the post gap material, there is a significant difference when wh-licensing interaction is measured in the post-gap material ($p<0.05$ for the \textit{object}--\textit{adj-front} difference; $p<0.01$ for the \textit{object}--\textit{adj-back} difference). Results are also significant when the whole embedded clause is measured ($p<0.01$ for both differences). To sum up: In all cases, the placement of a gap within an adjunct results in a significantly lower licensing interaction. This difference in licensing interaction suggests that the models have learned the AC inasmuch as they have attenuated expectations for wh-licensing within sentential adjuncts.

\subsection{Complex NP and Subject Islands}

The \key{Complex NP Constraint} (CNPC) holds that a gap cannot be hosted in a sentential clause dominated by a noun phrase with a lexical head noun. This constraint accounts for the unacceptability of \ref{ex:cnpcobj-rc-that}, \ref{ex:cnpcobj-rc-wh}, \ref{ex:cnpcsubj-rc-that} and \ref{ex:cnpcsubj-rc-wh} below. The CNPC does not apply to other NP modifiers, such as PPs, unless the modified NP occurs in subject position \citep{huang1982logical}. This ban, called the \key{Subject Constraint} (SC), accounts for the unacceptability of \ref{ex:cnpcsubj-prep} compared to \ref{ex:cnpcobj-prep}.

\ex. \label{ex:cnpc-islands}
\small{}
\a. \label{ex:cnpcobj-obj}
	\def\arraystretch{1.1}
	\begin{tabular}{ p{6.5cm} }
    \textit{object} \\ 
    \textbf{I know what the family bought \_\_  last year.}\\
	\small{Extraction of embedded clause object.} \\
	\end{tabular}
\b. \label{ex:cnpcobj-rc-that}
	\begin{tabular}{ p{6.5cm} } 
    \hline
	\textit{that-rc/obj} \\ 
	\textbf{*I know who the family bought the painting that depicted \_\_  last year.}\\
	\small{Extraction from `that'-headed relative clause modifying embedded object. CNPC violation.} \\
	\end{tabular}
\c. \label{ex:cnpcobj-rc-wh}
	\begin{tabular}{ p{6.5cm} } 
    \hline
	\textit{wh-rc/obj} \\
	\textbf{*I know who the family bought the painting which depicted \_\_  last year.}\\ 
	\small{Extraction from `wh'-headed relative clause modifying embedded object. CNPC violation} \\
	\end{tabular}
\d. \label{ex:cnpcobj-prep}
	\def\arraystretch{1.1}
	\begin{tabular}{ p{6.5cm} }
    \hline
    \textit{prep/obj} \\ 
    \textbf{I know who the family bought the painting by \_\_ last year.}\\
	\small{Extraction from PP attached to embedded object.} \\
	\end{tabular}
\e. \label{ex:cnpcsubj-subj}
	\begin{tabular}{ p{6.5cm} } 
    \hline
	\textit{subject} \\ 
	\textbf{I know what \_\_ fetched a high price at auction.} \\
	\small{Extraction of embedded clause subject.} \\
	\end{tabular}
\f. \label{ex:cnpcsubj-rc-that}
	\begin{tabular}{ p{6.5cm} } 
    \hline
	\textit{that-rc/subj} \\
	\textbf{*I know who the painting that depicted \_\_ fetched a high price at auction.}\\ 
	\small{Extraction from `that'-headed relative clause modifying embedded subject. CNPC violation} \\
	\end{tabular}
\f. \label{ex:cnpcsubj-rc-wh}
	\def\arraystretch{1.1}
	\begin{tabular}{ p{6.5cm} }
    \hline
    \textit{wh-rc/subj} \\ 
    \textbf{*I know who the painting which depicted \_\_ fetched a high price at auction.}\\
	\small{Extraction from `wh'-headed relative clause modifying embedded subject. CNPC violation.} \\
	\end{tabular}
\f. \label{ex:cnpcsubj-prep}
	\begin{tabular}{ p{6.5cm} } 
    \hline
	\textit{prep/subj} \\ 
	\textbf{*I know who the painting by \_\_ fetched a high price at auction.}\\
	\small{Extraction from PP attached to embedded subject.  SC  violation.} \\
	\end{tabular}

To test whether RNNs were sensitive to the CNPC and SC, we constructed 21 items for the variants shown in \ref{ex:cnpc-islands}, which resulted in 8 conditions. For \textit{prep/obj} and \textit{prep/subj} special care was taken to use prepositions that unambiguously attach to the object and subject NP, respectively. As post gap material varied between variants, only whole-clause wh-licensing interaction measurement is given for this experiment.  

\begin{figure}
\begin{minipage}{0.22\textwidth}
\includegraphics[width=\textwidth]{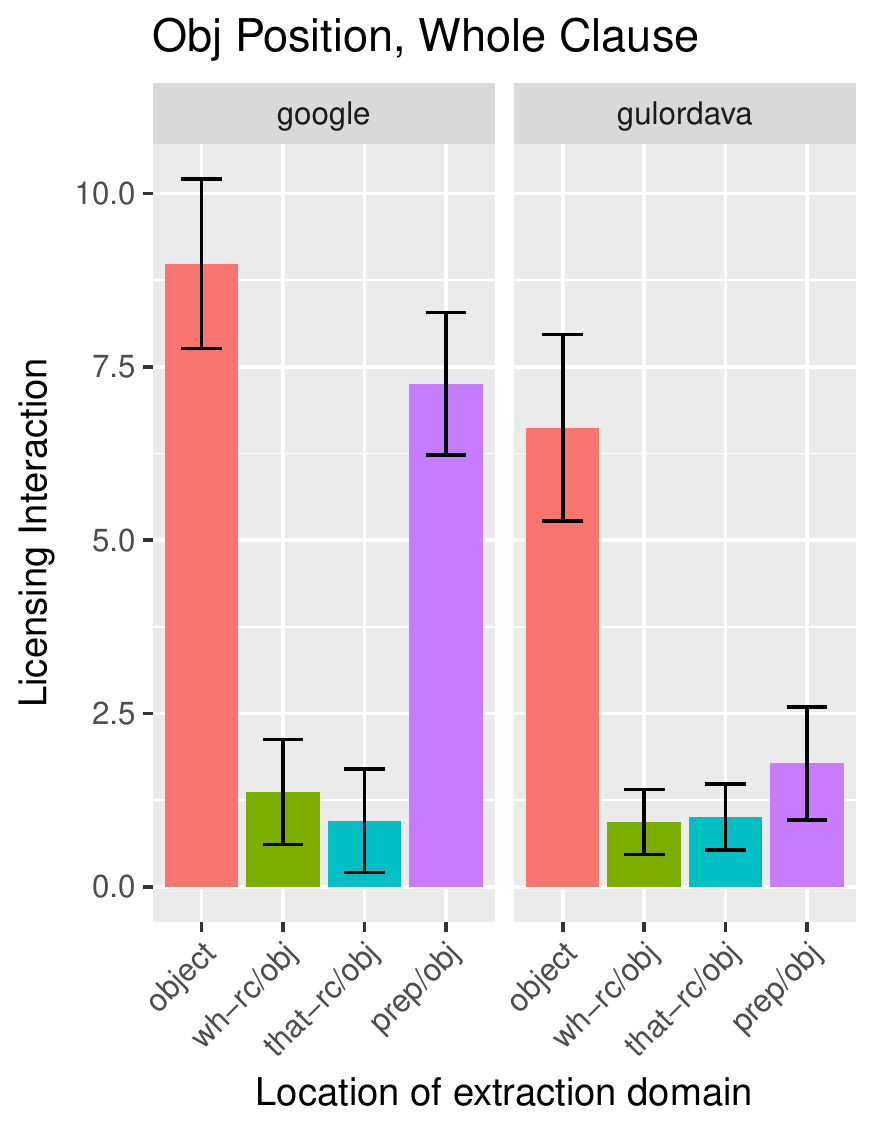}
\end{minipage}
\begin{minipage}{0.22\textwidth}
\includegraphics[width=\textwidth]{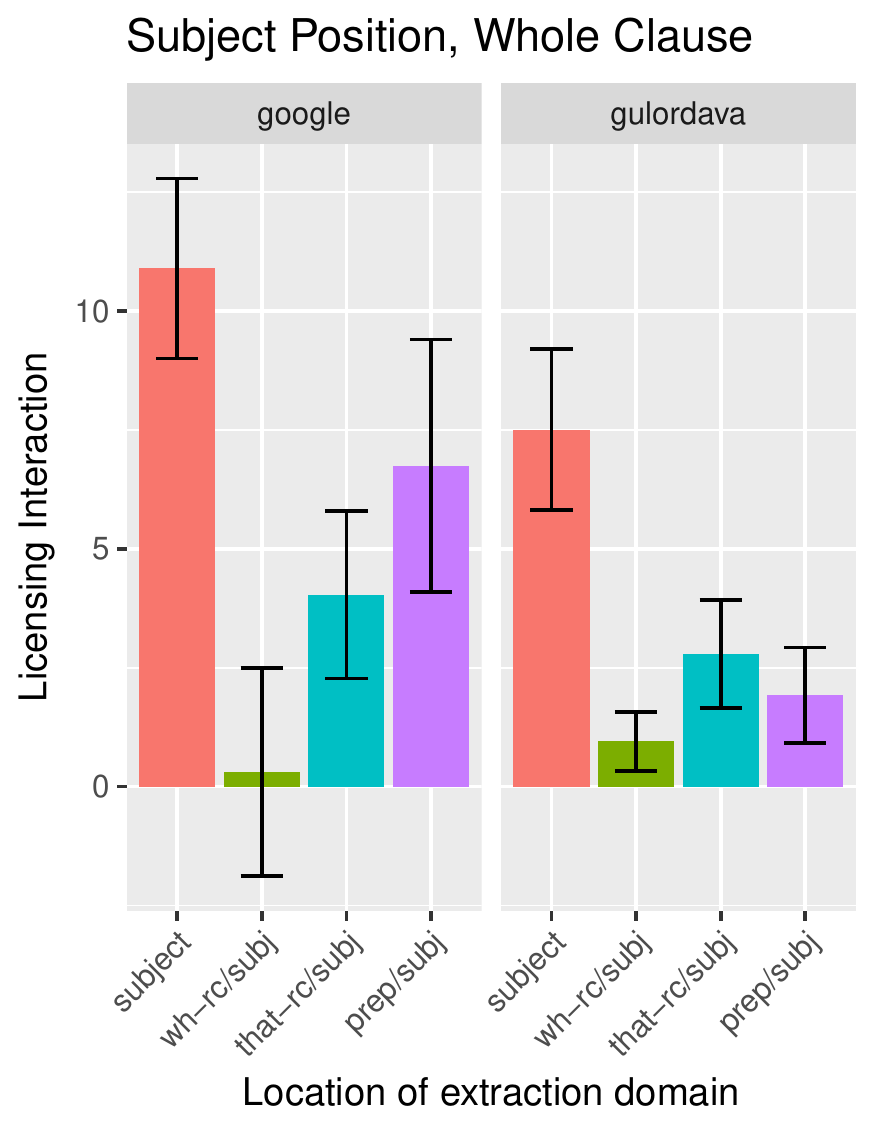}
\end{minipage}
\caption{Effect of extraction site location in complex np islands on wh-licensing interaction, measurement taken across the whole embedded clause. Object position is at left, subject position at right.}
\label{fig:cnpc-results}
\end{figure}

Results for object variants can be seen in the left panel of Figure \ref{fig:cnpc-results}, and results for the subject variants on the right. In all cases the comparatively large licensing interaction in non-island conditions (\emph{object} and \emph{subject}) shrinks when the extracted material occurs inside a complex NP (the middle bars in each chart). For the Google model the difference is significant for both CNP islands when extraction occurs in object position ($p<0.001$). For subject position, the difference is significant when the RC is headed by a wh-word (\textit{wh-rc/subj}) ($p<0.05$), but there is no significant difference when the RC is headed by `that', as in \textit{wh-that/subj}. For the Gulordava model, both differences are significant in subject ($p<0.05$) and object position ($p<0.01$). Of the eight comparisons in \ref{fig:cnpc-results} between CNPC islands and their non-island counterparts, seven show significant reduction in wh-licensing interaction. These differences indicate that both LSTMs do not generally expect extraction to occur from within complex NPs. 

However, the LSTMs demonstrate divergent licensing behavior when extraction occurs from out of a prepositional phrase. If the models were learning the SC, we would expect no significant difference between \emph{object} and \emph{prep/obj}, but a island-like reduction in licensing interaction between the \emph{subject} and \emph{prep/subj} conditions. However, for the Google model there is no significant difference in licensing interaction in any condition, and for the Gulordava model the difference is significant ($p < 0.05$) in all cases. These results demonstrate that neither model has learned the subject constraint, categorizing PPs as either licit extraction domains in all positions (the Google model) or treating them like islands (the Gulordava model).

\section{Conclusion}
\label{sec:conclusion}

We have provided evidence that state-of-the-art LSTM language models have learned to represent filler--gap dependencies and some of the constraints on them. These results capture the bi-directional nature of the dependency, due to the fact that our measure---wh-licensing interaction---measures both the salutary effect of a gap given the presence of an upstream filler, as well as the salutary effect of a filler given a gap. We found strong licensing effects in both subject, object and indirect object locations, as well as an expectation that the filler--gap relationship was one-to-one and relatively unaffected by grammatically-irrelevant interveners. The models also learned constraints on the dependency, insofar as licensing effect shrank when gaps were located in wh-islands, adjunct islands and most complex NP islands, although the subject constraint was not clearly learned and some trace licensing interaction remained.

While the Google model was trained on ten times more data, contained ten times as many hidden units and uses character CNN embeddings, its performance was not qualitatively more human-like than the Gulordava model. Both models failed to correctly generalize island constraints in two conditions: The Google model failed to learn that-headed Complex-NP Islands, the Gulordava model to learn Wh-Islands, and both failed to learn Subject Islands. These results indicate that---beyond a certain point---increased model size and training regimen give diminishing returns.

In other recent work, \citet{chowdhury2018rnn} tested the ability of neural networks to separate grammatical from ungrammatical extractions using similar metrics to ours, finding that their neural networks do not represent the unboundedness of filler--gap dependencies nor certain strong island constraints. We believe the difference between our results and theirs is due to experimental design: They choose to measure the probability of the question mark punctuation as a proxy for the RNNs gap expectation, and use sentence schemata instead of hand-engineered experimental items. While \citet{chowdhury2018rnn} conclude that the networks are not learning island-like constraints, but rather displaying sensitivity to syntactic complexity plus order, we demonstrate island-like effects where both the island and the non-island item are equally complex (in e.g. wh-islands). Note also that our work is focused on finding evidence that networks represent the probabilistic contingencies implied by island constraints, without attempting to directly model grammaticality judgments.

Our work shows these dependencies and their constraints can be learned to some extent by a generic sequence model with no obvious inductive bias for hierarchical structures. This is evidence against the idea that such an inductive bias is necessary for language learning, although the amount of data these models are trained on is much larger than the typical input to a child learner. 


\section*{Acknowledgements}
\small All experimental materials and scripts are available at \url{https://osf.io/zpfxm/}. EGW would like to acknowledge support from the Mind Brain Behavior Graduate Student Grant, as well as Emmanuel Dupoux and the Cognitive Machine Learning Group at the ENS. RPL gratefully acknowledges support to his laboratory from Elemental Cognition and from the MIT-IBM Watson AI Lab. This work was supported by a GPU Grant from the NVIDIA corporation.

\bibliographystyle{acl}
\bibliography{everything}

\end{document}